\documentclass[letterpaper, 10 pt, conference]{ieeeconf} 
\IEEEoverridecommandlockouts
\pdfoutput=1 
\let\labelindent\relax         
\newcommand{\figurename}{Fig.}

\usepackage{amssymb}
\usepackage{mathtools}

\mathchardef\mhyphen="2D   

\newcommand{\R}     {\mathbb{R}}          
\newcommand{\T}     {\top}                
\newcommand{\I}     {\mathbf{I}}          
\renewcommand{\P}     {\mathbf{P}}          

\newcommand{\pinv}      [1]{{#1}^+}
\newcommand{\vectornorm}[1]{||#1||}
\newcommand{\abs}	[1]{|#1|}

\newcommand{\nd}      {n}                              
\newcommand{\Nd}      {\mathcal{\MakeUppercase{\nd}}}  

\newcommand{\bx}     {\mathbf{x}}         

\newcommand{\dimQ}    {\mathcal{Q}}            
\newcommand{\dimK}    {\mathcal{K}}            

\newcommand{\bA}     {\mathbf{A}}         
\newcommand{\bN}     {\mathbf{N}}         
\newcommand{\bq}     {\mathbf{q}}         
\newcommand{\br}     {\mathbf{r}}         
\newcommand{\bqdot}  {\dot{\bq}}          

\newcommand{\bF}     {\mathbf{F}}         
\newcommand{\bqddot} {\ddot{\bq}}         
\newcommand{\btau}   {\boldsymbol{\tau}}  

\newcommand{\bM}{\mathbf{M}}
\newcommand{\bh }{\mathbf{h}}

\renewcommand{\nu}  {q}                   
\newcommand  {\dimU}{\mathcal{\MakeUppercase{\nu}}} 
\newcommand  {\bpi} {\boldsymbol{\pi}}    

\newcommand{\dimX}{\mathcal{\MakeUppercase{P}}}
\newcommand{\TaskPos}	{\mathbf x}

\newcommand{\TaskAcc}	{\ddot{\TaskPos}}

\newcommand{\Jacobian}	{\mathbf J}

\newcommand{\bLambda}     {\boldsymbol{\Lambda}}       %

\usepackage{textcomp} 

\newcommand {\bK} {\mathbf{K}}
\newcommand {\bD} {\mathbf{D}}
\newcommand {\ErrTaskPos}{\tilde{\TaskPos}}
\newcommand {\ErrTaskVel}{\dot{\ErrTaskPos}}
\newcommand {\ErrTaskAcc}{\ddot{\ErrTaskPos}}

\renewcommand	{\nu}   {q}                      
\renewcommand 	{\nd} 	{d} 			

\DeclareMathOperator{\bZero }{\mathbf{0}} 
\newcommand{\graspMatrix}{\mathbf{G}}

\newcommand{\TaskForce} {\mathbf{F}}

\newcommand{\ExternalForce}{\mathbf{F}_x}

\newcommand	{\ContactForce}		{\boldsymbol{\lambda}_c}
\newcommand	{\contactForce}[1] 	{\boldsymbol{\lambda}_{f#1}}
\newcommand 	{\contactMoment}[1]	{\boldsymbol{\lambda}_{m#1}}

\newcommand	{\contactForceElement}[1]{\lambda_{f,#1}}
\newcommand	{\contactMomentElement}[1]{\lambda_{m,#1}}

\newcommand	{\contactForceTangent}[1]{\lambda_{f,#1}}
\newcommand 	{\moment}[1] 		{\mathbf{m}_{#1}}
\newcommand 	{\momentNormal}[1] 	{\moment{z}}

\newcommand	{\skewMatrix}[1]	{\mathbb{S}(#1)}

\newcommand	{\externalWrench}	{\eta}
\newcommand 	{\commandWrench} 	{F}
\newcommand	{\ExternalWrench}	{\boldsymbol{\externalWrench}}
\newcommand 	{\CommandWrench} 	{\mathbf{\commandWrench}_c}
\newcommand	{\externalForceElement}	[1]{\eta_{f,#1}}
\newcommand	{\externalMomentElement}[1]{\eta_{m,#1}}
\newcommand 	{\commandForceElement}[1] 	{F_{f,#1}}
\newcommand 	{\commandMomentElement}[1] 	{F_{m,#1}}

\newcommand {\radius} {r}
\newcommand {\speed} {s}

\newcommand{\dx}{\delta x}
\newcommand{\dy}{\delta y}       
\usepackage{graphicx}
\graphicspath{{fig/}} 

\newcommand*{\sref}[1]{\S\ref{s:#1}}            
\newcommand*{\fref}[1]{\figurename~\ref{f:#1}}  
\newcommand*{\eref}[1]{(\ref{e:#1})}            

\usepackage[inline]{enumitem}
\setlist{nolistsep}

\usepackage{times}
\usepackage{graphicx}\graphicspath{{fig/}} 
\usepackage{epstopdf}
\usepackage{algorithm}
\usepackage{algorithmic}
\usepackage{multirow} 
\usepackage{multicol}
\usepackage{array}
\usepackage{mdwmath}
\usepackage{mdwtab}
\usepackage{rotating}
\usepackage{caption}
\usepackage{subfig}
\usepackage{amssymb}
\usepackage{verbatim}
\usepackage{stfloats}
\usepackage{color}
\usepackage[normalem]{ulem}                                        
\usepackage[percent]{overpic} 
\makeatletter\newcommand{\manuallabel}[2]{\def\@currentlabel{#2}\label{#1}}\makeatother
\usepackage[disable]{todonotes}{}{}\newcommand{\maybedo}[1]{}

\definecolor{lightgrey}{RGB}{155,155,155}

\definecolor{orange}{RGB}{255,128,0}

\usepackage{amsmath}
\usepackage{nomencl}
\makenomenclature

\makeatletter\newcommand{\mylabel}[2]{\def\@currentlabel{#2}\label{#1}}\makeatother

\title{\LARGE\bf A Projected Inverse Dynamics Approach for Dual-arm Cartesian Impedance Control}

\author{
  Hsiu-Chin Lin, Joshua Smith, Keyhan Kouhkiloui Babarahmati, Niels Dehio, and Michael Mistry
  \thanks{
    H. Lin ({\tt\small h.lin@bham.ac.uk}), Joshua Smith, and Michael Mistry are at the School of Informatics, University of Edinburgh, UK.
    Keyhan Kouhkiloui Babarahmati is at the School of Computer Science, University of Birmingham, UK.
    Niels Dehio is at the Research Institute for Robotics and Process Control, Technical University Braunschweig, Germany}
}
\renewcommand{\baselinestretch}{.96}\selectfont

\begin{document}

\listoftodos\clearpage
\setlength{\floatsep}{5pt}          
\setlength{\textfloatsep}{7pt}      
\maketitle
\thispagestyle{empty}
\pagestyle{empty}

\begin{abstract}
  \noindent We propose a method for dual-arm manipulation of rigid objects, subject to external disturbance. The problem is formulated as a Cartesian impedance controller within a projected inverse dynamics framework. We use the constrained component of the controller to enforce contact and the unconstrained controller to accomplish the task with a desired 6-DOF impedance behaviour.  Furthermore, the proposed method optimises the torque required to maintain contact, subject to unknown disturbances, and can do so without direct measurement of external force. The techniques are evaluated on a single-arm wiping a table and a dual-arm platform manipulating a rigid object of unknown mass and with human interaction.
  \cite{IROS.2017}
\end{abstract}

\section{Introduction}  \label{s:introduction}      	
\noindent Many activities in robotics can be described in terms of performing a desired task subject to physical \emph{constraints} and external \emph{disturbances}. For example, a dual-arm robot squeezing a rigid object or sliding it on a table (i.e.\ constraints), while a human interacts by pushing the object or adding unknown mass (i.e.\ disturbances) (\fref{intro}). The distinction between the two forces is critical as constraint forces do not induce motion, while disturbance forces can.  A controller must be aware of contributions from both types of forces in order to achieve its task in an optimal manner. For example, to counteract disturbances with a desired impedance response, while squeezing only as necessary to maintain contact of the object.


In this paper, we propose a method for dual-arm manipulation of a rigid object subject to external disturbance. The problem is formulated in the projected inverse dynamics framework~\cite{Aghili.2005}, such that the constrained component of the controller enforces the contact and the unconstrained controller accomplishes the task and a desired Cartesian impedance behaviour. The proposed method also attempts to apply the minimum torque required to maintain the contact, and does so without use of Force/Torque sensing at the contacts. The techniques are evaluated on both single-arm (sliding on a table) and dual-arm (manipulating a rigid object) platforms with various examples. 

\begin{figure}[t!]\centering%
 \includegraphics[width=.8\linewidth]{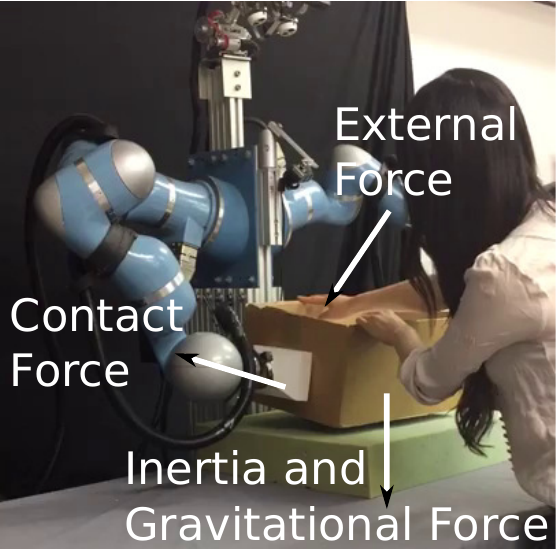}\mylabel{f:dual-conceptual}{(top)}
  \caption{A dual-arm robot manipulates an object while a human is interacting by pushing or pulling the object}
  \label{f:intro}%
\end{figure}

\section{Background} 	\label{s:problem_definition}	\noindent Our work stems from prior literature in Dual- (or Multi-) Arm Manipulation, Impedance Control, Grasping, Projected Dynamics and Operational Space Control. A pioneering work on multi-arm manipulation is given in~\cite{Hayati.1986}. Depending on the type of interaction, there are two categories of cooperative manipulator strategies: constraining the \emph{relative motion} of two arms~\cite{Chiaverini.1996}, or controlling the \emph{internal force}~\cite{Bonitz.1996}\cite{Caccavale.2008}\cite{Hirche.2015}. The former constrains the relative motion between two end-effectors but allows rolling of the object, while the later category attempts to control internal force, i.e.\ forces producing no motion upon the object.

Maintenance of contact when grasping requires dealing with unknown forces and moments applied to the object, which may include disturbances arising from motion of the robot, inertial forces during manipulation, or the forces due to gravity. For this, previous work seeks contact forces that prevent  separation or sliding of the contacts, either for locomotion~\cite{Righetti.2013}\cite{Lee.20016} or grasping~\cite{Bonitz.1996}\cite{Bicchi.2000}, and several are formed into constraint optimisation problems~\cite{Kerr1986-rh}\cite{Buss.1996}\cite{Trinkle.1997}. 

The classical impedance controller was introduced to deal with interaction~\cite{Hogan.1985}, and was later extended for the impedance control of an object held by two manipulators~\cite{Schneider.1992}. The work of~\cite{Caccavale.2008} further extends this paradigm to realise a 6-DOF desired object impedance from two manipulators, while inducing an internal impedance controller to maintain contact without large forces. 

In most manipulation problems, the desired task is specified in an end-effector reference frame. Operational space control~\cite{1987.IJRR.Khatib} decomposes rigid-body dynamics into task-relevant and irrelevant subspaces. For systems with physical constraints,~\cite{Aghili.2005} uses orthogonal decomposition to decompose dynamics into constrained and unconstrained subspaces. Both were extended by~\cite{Mistry.2011} to enable constraints within the operational space formulation.

Our main control framework is based on extending the previous work of \cite{Mistry.2011} from single arm to dual-arm, and applying it to achieve 6-DOF impedance of a rigid object in contact. Furthermore, we are able to optimise the torque required to maintain the grasp, while subject to unknown disturbances, unknown mass of the object, and without use of Force/Torque sensing at the contact points. 
\section{Method}        \label{s:method}    		\noindent 
The projected inverse dynamics formulation was originally used to control single arm acting on a rigid environment. In this work, it is extended to multiple arms manipulating a single rigid object, together with 6-DOF Cartesian impedance controller at the object and optimisation of contact forces.

\subsection{Projected Inverse Dynamics}

\noindent Let $\bq, \bqdot, \bqddot \in \R^\dimQ$ denote the joint positions, velocities, and accelerations of a $\dimQ$ degree-of-freedom manipulator, the dynamics can be expressed in the Lagrangian form
\begin{equation}
 \bM \bqddot + \bh = \btau
\end{equation}
where $\btau \in \R^\dimU$ is the vector of joint torques, $\bM \in \R^{\dimU\times \dimU}$ is the inertia matrix, and $\bh \in \R^{\dimU}$ is the vector of centrifugal, gyroscopic, and Coriolis effects, and generalised gravitational torque.

When a robot is interacting with the environment, the end-effector motion may be subject to the constraints imposed by the environment, which modifies the motion in order to accommodate the constraints. An additional term is added to describe the rigid body dynamics under constraints 
\begin{equation}
  \bM \bqddot + \bh = \btau + \Jacobian_c^\T \ContactForce
  \label{e:dynamics-contact}
\end{equation}
where $\Jacobian_c \in \R^{\dimK \times \dimQ}$ is the constraint Jacobian that describes $\dimK$ linearly independent constraints, and $\ContactForce$ are the constraint forces due to contact that enforce the following conditions: 
\begin{equation}
  \begin{aligned}
    & \Jacobian_c\bqdot = \bZero \\ 
    & \Jacobian_c\bqddot + \dot{\Jacobian}_c\bqdot = \bZero.    
  \end{aligned}
  \label{e:constraint-jc}
\end{equation}

\noindent 
\cite{Aghili.2005}  proposed the projected inverse dynamics framework, such that the dynamics equation in \eref{dynamics-contact} may be decomposed into constrained and unconstrained components;
\begin{equation}
   \btau = \P \btau + (\I - \P ) \btau
   \label{e:dynamics-decomposition}
\end{equation}
where $\P =\I - \pinv{\Jacobian}_c \Jacobian_c$ is the orthogonal projection matrix that projects arbitrary vectors into the null space of the constraint Jacobian $\Jacobian_c$ and $\pinv{\Jacobian}_c$ is the Moore-Penrose pseudo-inverse of $\Jacobian_c$. Note that the two terms in \eref{dynamics-decomposition} are orthogonal to each other $\P\btau \perp (\I-\P) \btau$ such that the first term $\P \btau = \P \bM \bqddot + \P \bh$ generates no motion in the constraint space, and the second term $(\I-\P) \btau$ enforces the constraint without generating joint motion. 

\cite{1987.IJRR.Khatib} introduced the {\em operational-space} formulation to address the dynamics of task-space movement:
\begin{equation}
 \bF = \bLambda \TaskAcc + \bLambda \Jacobian_x \bM^{-1} \bh - \dot{\Jacobian}_x \bqdot
\end{equation}
where $\bF$ is the force applied at the end-effector for the desired acceleration $\TaskAcc$, $\Jacobian_x$ is the Jacobian at $\TaskPos \in SE(3)$ , and $\bLambda= (\Jacobian_x\bM^{-1} \Jacobian_x^\T)^{-1}$ is the operational space inertia matrix. \cite{Mistry.2011} proposed operational space controllers for constrained dynamical systems such that the term $\P \btau $ in \eref{dynamics-decomposition} is replaced by $\P\Jacobian_x^\T\TaskForce$, and $\TaskForce$ is the force applied at the end-effector for the desired acceleration $\TaskAcc$:
\begin{equation}
 \bF = \bLambda_c \TaskAcc + \bLambda_c \left[ \Jacobian_x \bM_c^{-1} (\P\bh -\dot{\P}\bqdot )- \dot{\Jacobian}_x \bqdot \right]
\end{equation}
where $\bLambda_c=(\Jacobian_x\bM_c^{-1} \P \Jacobian_x^\T)^{-1}$ and $\bM_c=\P\bM+\I-\P$ are the constraint consistent operational space and joint space inertia matrix, respectively.

\subsection{Impedance Controller}
\label{s:method-impedance}
\noindent 
The objective of Cartesian impedance control is to dictate the disturbance response of the robot, at a particular contact location. If a given operational location $\bx \in SE(3)$ is subject to an external disturbance $\ExternalForce$, we would like the resulting motion to be prescribed as 
\begin{equation}
  \bLambda_d \ErrTaskAcc + \bD_d \ErrTaskVel + \bK_d \ErrTaskPos = \ExternalForce 
  \label{e:external-force}
\end{equation}
where $\ErrTaskPos=\TaskPos- \TaskPos_d$ and $\TaskPos_d$ is a virtual equilibrium point, $\bLambda_d$, $\bD_d$, and $\bK_d$ are desired inertia, damping, and stiffness matrices, respectively. 

Adding the external disturbance force into the inverse dynamics equation \eref{dynamics-contact}, the general rigid-body dynamics can be described as 
\begin{equation}
  \bM \bqddot + \bh = \btau + \Jacobian_c^\T \ContactForce + \Jacobian_x^\T \ExternalForce
  \label{e:dynamics-external}
\end{equation}
  
\noindent As only the unconstrained component of the control torque $\P \btau$ contributes to motion (and thus the desired impedance behaviour), we multiply both sides of \eref{dynamics-contact} by $\P$, eliminating constraint forces, and resulting in 
\begin{equation}
  \P \bM \bqddot + \P \bh =\P (\btau + \Jacobian_x^\T \ExternalForce).
  \label{e:dynamics-unconstrained}
\end{equation}

\noindent
Writing \eref{dynamics-unconstrained} in operational space yields (see Appendix):
\begin{equation}
   \TaskForce + \ExternalForce = \bh_c + \bLambda_c \TaskAcc,
  \label{e:impedance-controller}
\end{equation}
where  $\TaskForce$ is force due to actuation (control) and $\ExternalForce$ is the external disturbance. 

From the classical impedance controller \cite{Hogan.1985,Ott.2008}, the control input $\TaskForce$ which leads to the desired impedance behaviour is given by
\begin{equation}
  \begin{aligned}
  \TaskForce & = \bh_c + \bLambda_c \TaskAcc_d - \bLambda_c \bLambda_d^{-1} ( \bD_d \ErrTaskVel + \bK_d \ErrTaskPos) \\ 
	    & + (\bLambda_c \bLambda_d^{-1} - \I) \ExternalForce   
  \end{aligned}
  \label{e:impedance-combined}
\end{equation}

\noindent 
If the desired inertia $\bLambda_d$ is identical to the robot inertia $\bLambda_c$, the feedback of the external force $\ExternalForce$ can be avoided. Substituting \eref{impedance-combined} into \eref{impedance-controller}, and multiply by $\bLambda_d \bLambda_c^{-1}$, and the control equation can be simplified to 
\begin{equation}
  \TaskForce = \bh_c + \bLambda_c \TaskAcc_d - \bD_d \ErrTaskVel - \bK_d \ErrTaskPos 
  \label{e:impedance-simplified}
\end{equation}
\noindent Using \eref{impedance-simplified}, the desired impedance response \eref{external-force} is achieved without measuring the external force.

An important insight is, in order to realise our desired impedance behaviour in \eref{external-force}, the control torque $\P\btau$ does not involve the constraint force. The constraint is enforced by
\begin{equation}
  (\I-\P) (\bM \bqddot + \bh) = (\I-\P) (\btau + \Jacobian_x^\T \ExternalForce) + \Jacobian_c^\T \ContactForce 
  \label{e:dynamic-constrained}
\end{equation}

\noindent The above equation generates constraint force without any effect on the joint space motion, or our desired impedance characteristic. We can exploit this property, for example, to compute the optimal control forces required to maintain grasp of an object. An overview of the control framework is illustrated in \fref{framework}
\begin{figure}
 \includegraphics[width=\linewidth]{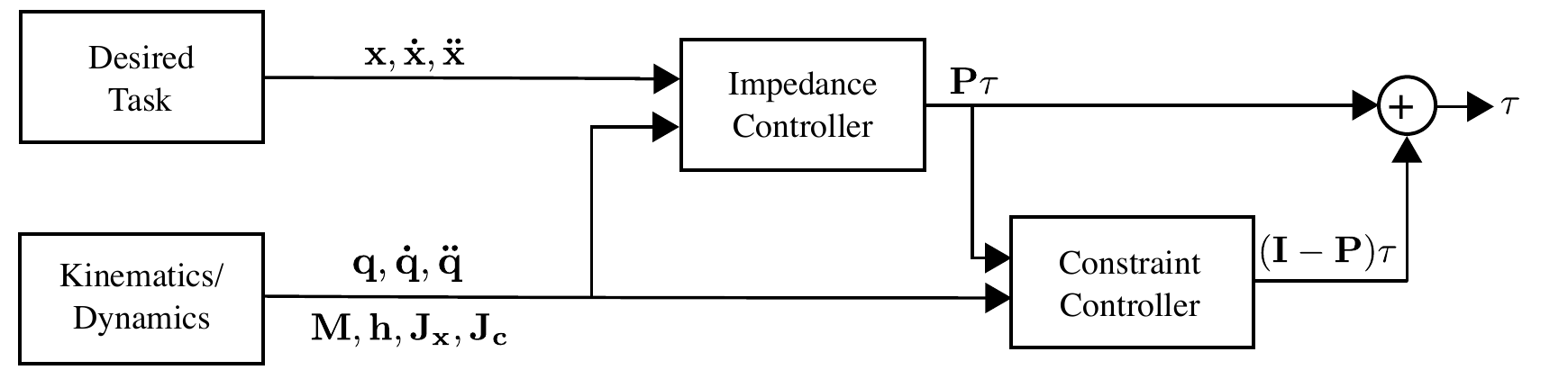}
 \caption{An overview of the projected inverse dynamics framework}
 \label{f:framework}
\end{figure}

\subsection{Multi-arm manipulation} 
\noindent 
For a multi-arm robot manipulating a single rigid object via a force-closed grasp, each end-effector is in contact with the object and may generate arbitrary wrenches upon the object (see \fref{contact-wrench}). The constraints are to enforce the force-closed grasp of the object and generate no motion that might violate the underlying task. For this, the constraint is formulated so that only {\em internal force} is allowed.

Based on the study in analytical dynamics~\cite{Udwadia.2001}, the constraint force does not produce any virtual work for any virtual displacement. From the analysis in~\cite{Hirche.2015}, {\em internal wrenches}, or end-effector wrench acting in the null space of grasp map, yields the same property with constraint force in a multi-arm system. For this, the dual-arm system is constrained such that only internal wrenches are allowed to enforce the contacts. 

Following the definition in~\cite{Murray.1994}, the grasp matrix of the $i^{th}$ arm in a multi-arm manipulation system is defined by the mapping between the object twist to the twists of the contacts (here written with respect to a common (global) coordinate frame): 
\[
 \graspMatrix_i \in\R^{6\times6} = \begin{bmatrix}
		  \I_{3\times 3} & \bZero_{3\times3} \\ 
		  \skewMatrix{\br_i} & \I_{3\times3}
                  \end{bmatrix}
\]
\noindent where $\br_i$ is relative distance from the contact position to the object centre-of-mass position, and $\skewMatrix{\br_i} \in\R^{3\times3}$ is the skew-symmetric matrix performing the cross product 
\[
  \skewMatrix{\br} = \begin{bmatrix} 0 & -r_z & r_y \\ r_z & 0 & -r_x \\ -r_y & r_x & 0 \end{bmatrix}
\]

\noindent 
Assuming that the robot has $\dimK$ manipulators, the grasp map $\graspMatrix$ is the horizontal concatenation of $\dimK$ grasp matrices. For example, the grasp map of the dual-arm system ($\dimK=2$) is defined as 
$\graspMatrix = \begin{bmatrix} 
\graspMatrix_L & \graspMatrix_R                 
                \end{bmatrix} \in \R^{6 \times 12}$ 
where $\graspMatrix_L, \graspMatrix_R$ are the grasp matrix of the left and the right arm.

The null space projection $\I- \pinv{\graspMatrix}  \graspMatrix$ projects any arbitrary vector onto the null space of the grasp map. The resulting contact force satisfies $\graspMatrix \ContactForce = 0$, yielding no net wrench on the object and contributing to only internal force. Under this formulation, the constraint Jacobian in \eref{dynamics-external} for a multi-arm system is written as:
\begin{equation}
  \Jacobian_c \in\R^{\dimK(\dimX \times \dimQ)} = (\I - \pinv{\graspMatrix } \graspMatrix)
		\begin{bmatrix}
		    \Jacobian_1 & & \bZero \\ 
				& \ddots &   \\
		         \bZero & & \Jacobian_{\dimK}
		\end{bmatrix} 
		\label{e:dual-jacobian}
\end{equation}
where $\dimX$ denotes the dimensionality of the end-effector space,  $\Jacobian_i \in \R^{\dimX \times \dimQ} $ are the Jacobian of the $i^{th}$ arm, and $\ContactForce \in \R^{\dimK\dimX}$ is the vertical concatenation of all contact wrenches due to internal forces that apply no net wrench on the object.

\subsection{Optimal Contact Wrenches}
\noindent
Maintenance of the contact requires dealing with unknown forces and moments applied to the object, which may include the disturbances arising from the motion of the robot, inertial forces during manipulation, or the forces due to gravity. For this, the contact wrench  applied by the hands should prevent the separation or sliding of the contact.

The contact wrench includes the contact force and the contact moment $\ContactForce\in\R^6=\left[\contactForce{}, \contactMoment{} \right]^\T$. Throughout the rest of this paper, we use the subscripts $f$ and $m$ to denote the force and moment, respectively, and we choose the z-axis as the direction normal to the contact surface. Specifically, the contact force $\contactForce{}=\left[\contactForceElement{x}, \contactForceElement{y}, \contactForceElement{z}\right]^\T$ where $\contactForceElement{z}$ is the normal force, and $\contactForceElement{x}, \contactForceElement{y}$ are the tangential forces. The moment $\contactMoment{} = \left[\contactMomentElement{x}, \contactMomentElement{y}, \contactMomentElement{z} \right]^\T$ are the moment along each axis. A dual-arm example is illustrated in \fref{contact-wrench}.

\begin{figure}[t!]
 \centering
  \includegraphics[width=\linewidth]{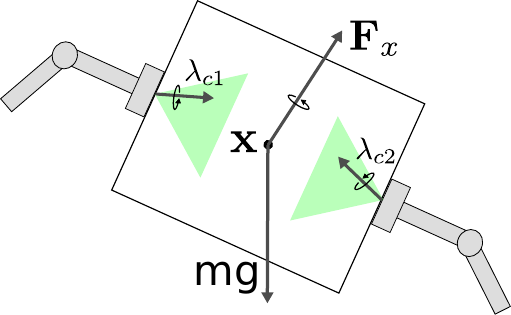}
  \caption{An illustration of a dual-arm manipulation problem. $\mathbf{F}_x$ is an external wrench applied at operational space point $\mathbf{x}$. Other forces acting on the object include inertial and gravitational forces (e.g.\ $\text{mg}$). The contact wrenches are $\lambda_{c1}$ and $\lambda_{c2}$, and their friction cones are illustrated in green. (For visualization purpose, we illustrate the contact forces and their friction cones in the opposite direction)}
  \label{f:contact-wrench} 
\end{figure}

\subsubsection{Unilateral Constraints}
The manipulators should only push toward the contact, but not pull, in order to maintain contacts.  Hence, the contact normal should satisfy the unilateral constraint
\begin{equation}
  \contactForceElement{z} \geq 0
  \label{e:constraints-unilateral}
\end{equation}

\subsubsection{Friction Cone Constraints}
If there is significant contact friction, a common way to describe the contact is by the Coulomb's friction model \cite{Trinkle.1997}. By Coulomb's Law, the magnitude of tangential force  $\contactForceTangent{}$ should not exceed the friction coefficient times the normal force  to avoid slipping
\begin{equation}  
  \mu \contactForceElement{z} \geq \sqrt{\contactForceElement{x}^2 + \contactForceElement{y}^2 } 
  \label{e:constraints-friction}
\end{equation}

\noindent where $\mu$ is the friction coefficient which depends on the material of the object. Geometrically, the set of forces which can be applied should lie in a cone centred about the direction normal to the contact surface (i.e.\ the grasp is more stable if the direction of the force is more orthogonal to the surface of the object). 

\subsubsection{Moment Constraints}
We assume the surface friction and the contact patch are large enough to generate friction force and moment. To avoid the hand from rolling at the contact point, the constraints are imposed on the torsional moment~\cite{Buss.1996} and shear moment~\cite{Bonitz.1996} 
\begin{equation} 
  \begin{aligned}
  \gamma \contactForceElement{z} & \geq \abs{\contactMomentElement{z} } \\
  \dx \contactForceElement{z} & \geq \abs{\contactMomentElement{x} } \\
  \dy \contactForceElement{z} & \geq \abs{\contactMomentElement{y} } 
  \label{e:constraints-moment}
  \end{aligned}
\end{equation}
where $\gamma$ is the torsional friction coefficient, and $\dx, \dy$ are the distance from the centre of the hand to the edge of the hand in $x$ and $y$ direction (assuming a rectangular contact patch). The latter two constraints ensure the contact centre of pressure remains within the contact patch of the hand. 

The optimal contact forces are the minimum torques needed to maintain all contacts while satisfying the unilateral, friction cone, and moment constraints at the contact points, \eref{constraints-unilateral} \eref{constraints-friction} \eref{constraints-moment} and balance out the external forces, including the forces acting on the object and the object dynamics. 

Assuming that we have $\dimK$ contacts ($\dimK=2$ for the dual arm example), there are $\dimK$ contact wrenches, and constraints for all the contact wrenches need to be solved. If the contact locations are fixed, finding the minimum torques is a convex optimisation problem over contact wrenches~\cite{Boyd.2007}.
\begin{equation}
 \begin{aligned}
  \underset{\btau} {\text{minimise   }} & (\I-\P) \btau \\
  \text{subject to } & \contactForceElement{z}^{i} \geq 0 \\
		    & \mu \contactForceElement{z}^{i} \geq \sqrt{ (\contactForceElement{x}^{i}) ^2 + (\contactForceElement{y}^{i})^2 } \\
		    & \gamma \contactForceElement{z}^{i} \geq \abs{\contactMomentElement{z}^{i} } \\
		    & \dx \contactForceElement{z}^{i} \geq \abs{\contactMomentElement{x}^{i} } \\
		    & \dy \contactForceElement{z}^{i} \geq \abs{\contactMomentElement{y}^{i} }   
  \end{aligned}
  \label{e:contact-force-optimisation}
\end{equation}

\noindent 
where the superscript $i$ denotes the $i^{th}$ contact. Following the methods proposed in \cite{Aghili.2016}, multiplying both sides in \eref{dynamic-constrained} by $\pinv{\Jacobian_c^\T}$, resulting a compact expression $\ContactForce = \ExternalWrench - \CommandWrench$ where $\CommandWrench= \pinv{\Jacobian_c^\T} (\I-\P) \btau$ is the equivalent end-effector wrench corresponding to the input torque $(\I-\P) \btau$, and $\ExternalWrench =\pinv{\Jacobian_c^\T}\left[(\I-\P) \bM \bM_c^{-1} ( \P \btau - \P \bh + \dot{\P} \bqdot + \Jacobian_x^\T\ExternalForce ) +(\I-\P)  \bh \right]$ can be interpreted as the sum of all external wrenches in the constrained space. Each element of the contact force can be described as 
\begin{equation}
  \begin{aligned}
    \contactForceElement{x}^{i} &= \externalForceElement{x}^{i} - \commandForceElement{x}^{i} \\
    \contactForceElement{y}^{i} &= \externalForceElement{y}^{i} - \commandForceElement{y}^{i} \\
    \contactForceElement{z}^{i} &= \externalForceElement{z}^{i} - \commandForceElement{z}^{i} \\
    &\vdots
  \end{aligned}
  \label{e:contact-force-decomposition}
\end{equation}

\noindent Substituting \eref{contact-force-decomposition} into \eref{contact-force-optimisation}, the constraint optimisation problem can be reformulated in terms of the unknown variable $\CommandWrench$:
\begin{equation}
 \begin{aligned}
  &\underset{\CommandWrench}{\text{minimise   }} \CommandWrench^\T \Jacobian_c \Jacobian_c^\T \CommandWrench \\
  &\text{subject to } \externalForceElement{z}^{i} - \commandForceElement{z}^{i}  \geq 0 \\
  & \mu ( \externalForceElement{z}^{i} - \commandForceElement{z}^{i} ) \geq \sqrt{( \externalForceElement{x}^{i} - \commandForceElement{x}^{i} )^2 + ( \externalForceElement{y}^{i} - \commandForceElement{y}^{i} )^2 } \\
    & \gamma 	( \externalForceElement{z}^{i} - \commandForceElement{z}^{i} ) \geq \abs{ \externalMomentElement{z}^{i} - \commandMomentElement{z}^{i} } \\
    & \dx 	( \externalForceElement{z}^{i} - \commandForceElement{z}^{i} ) \geq \abs{ \externalMomentElement{x}^{i} - \commandMomentElement{x}^{i} } \\
    & \dy 	( \externalForceElement{z}^{i} - \commandForceElement{z}^{i} ) \geq \abs{ \externalMomentElement{y}^{i} - \commandMomentElement{y}^{i} } \\
  \end{aligned}
  \label{e:commanded-force-optimisation}
\end{equation}

\noindent 
Note that \eref{contact-force-decomposition} and \eref{commanded-force-optimisation} require knowledge of the external disturbance $\ExternalForce$ (included in the $\ExternalWrench$ term). Remarkably, as our unconstrained controller maintains the desired impedance characteristic, we are able to estimate $\ExternalForce$ using the displacement of object from \eref{external-force}. Then, we arrive at an optimisation problem which seeks the commanded forces $\CommandWrench$ that require minimum torques to maintain the constraints without explicitly knowing the values of the contact forces.

Finally, as the friction cone constraints are quadratic (and therefore not realistic for real-time control) we approximate the constraints with linearised friction cones of 8 edges. The optimisation problem is then solved using GUROBI \cite{GUROBI} (quadratic program with linear constraints) and within the real-time control loop. 
      
\section{Evaluation}    \label{s:evaluation}        	\noindent %
We conduct experiments using our dual KUKA LWR platform Boris. Although the robots are equipped with force/torque sensors at the end-effector, these are only used for recording forces and not in the controller. 

\begin{figure*}[t!]%
  \centering
   \begin{overpic} [width=.24\linewidth]{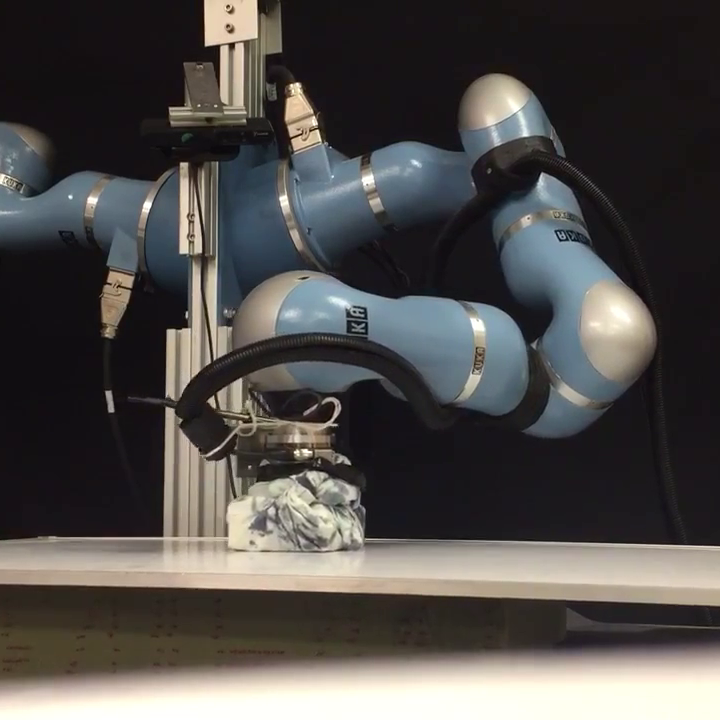}	{\manuallabel{f:single-static}{(a)}\color{black}\ref{f:single-static}} 	\end{overpic}~
   \begin{overpic} [width=.24\linewidth]{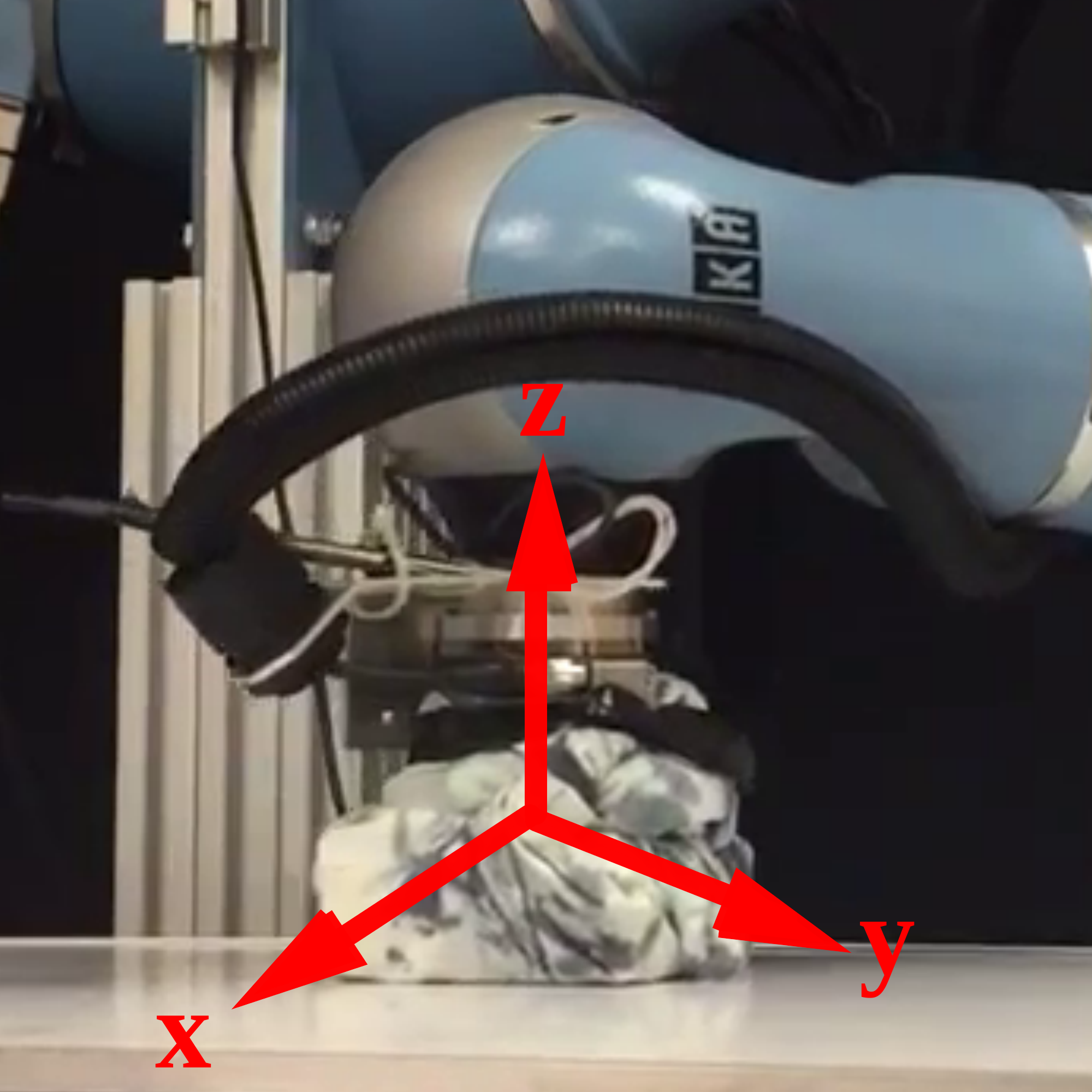}	{\manuallabel{f:single-coord}{(b)} \color{black}\ref{f:single-coord}}	\end{overpic}~
   \begin{overpic} [width=.24\linewidth]{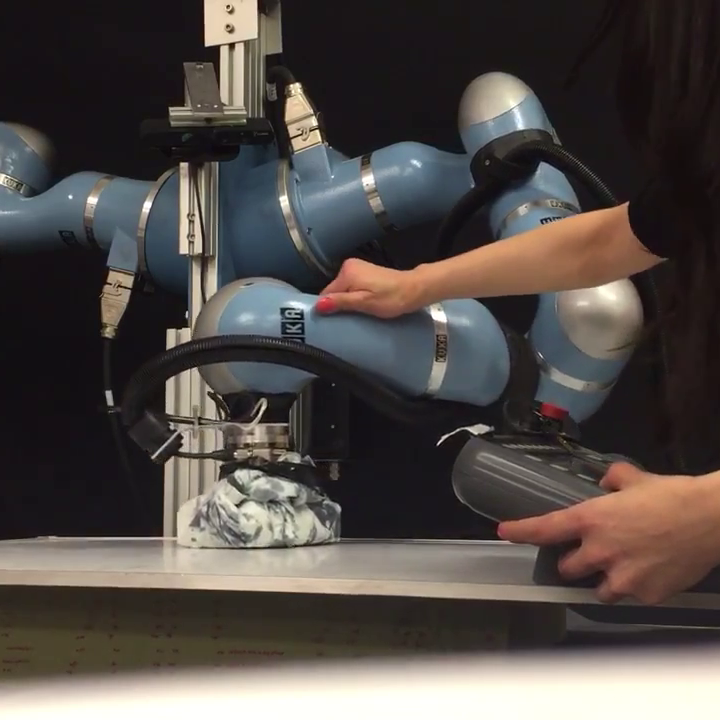}	{\manuallabel{f:single-human}{(c)} \color{black}\ref{f:single-human}}	\end{overpic}~
   \begin{overpic} [width=.24\linewidth]{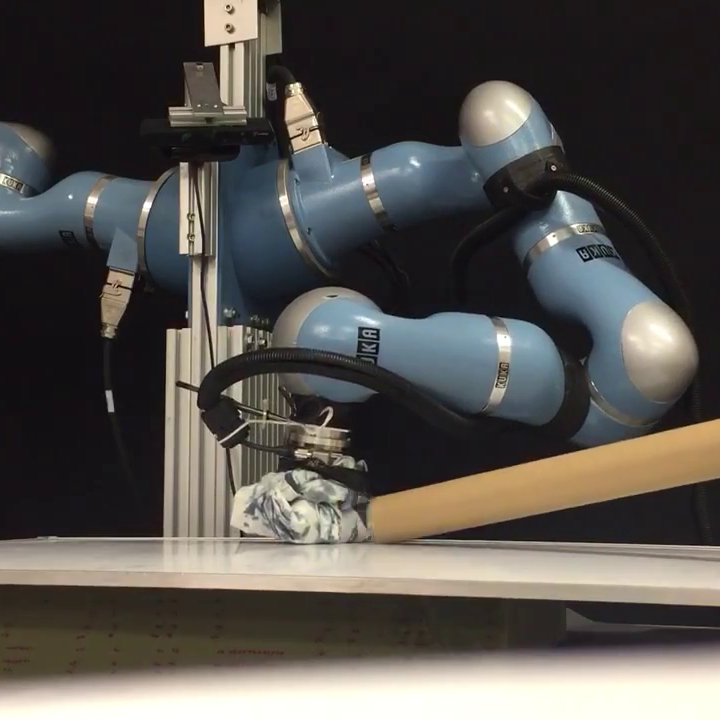}	{\manuallabel{f:single-pokes}{(d)}\color{black}\ref{f:single-pokes}} 	\end{overpic}~   
 \\\vspace{-1mm} 
  \caption{Experiment of a single arm robot. The end-effector moving in a circular motion on the table surface \ref{f:single-static} and the constraint on z direction and the xy moment \ref{f:single-coord}. While the robot is performing the circular motion, a human moves the hand to a different position \ref{f:single-human} and pokes the robot with a a stick \ref{f:single-pokes}.}
  \label{f:single-exp} 
\end{figure*}  

\subsection{Single arm wiping}
\noindent For a proof of concept on a simpler problem, the proposed approach is first applied on a single manipulator (see \fref{single-exp}\ref{f:single-static}) in contact with a flat rigid surface. The aim is to perform a Cartesian wiping task on the surface while maintaining a desired impedance characteristic and contact with minimal torques. 

To ensure the end-effector stays in contact with the table, constraints are imposed on the vertical position of the hand $z$ as well as the two roll directions $x,y$ (see \fref{single-exp}\ref{f:single-coord}).  For this, the constraint is described by $\Jacobian_c \in \R^{3\times 7}$ that relates joint space to $z$ position and rotations along the $x$ and $y$ axes.

The robot is resting on the table at the beginning of the experiment, and the task is to follow a circular trajectory on the $x,y$-plane and maintain a fixed rotation along the direction normal to the contact surface. The desired end-effector trajectory is described as
$\bx_d \in\R^3 = \begin{bmatrix} \radius \cos(\speed t),  \radius \sin(\speed t), 0 \end{bmatrix}^\T$ where $\radius$ is the radius of the circular motion, $\speed$ is a parameter to control the speed, $t$ is the time difference from the beginning of the circular motion, and $0$ is the desired yaw angle.

In \fref{single-circular} \ref{f:single-commanded}, the magnitude of commanded force at the end-effector is plotted from a fraction of the data collected. We can see that the commanded force varies depending on the configuration of the robot. When the hand accelerates along its shorter side, the hand is more prone to rolling ($\delta y < \delta x$ in the constraints defined by \eref{constraints-moment}, and consequently, the robot pushes harder.
\begin{figure}[t!]
    \centering  
    \includegraphics[width=.95\linewidth]{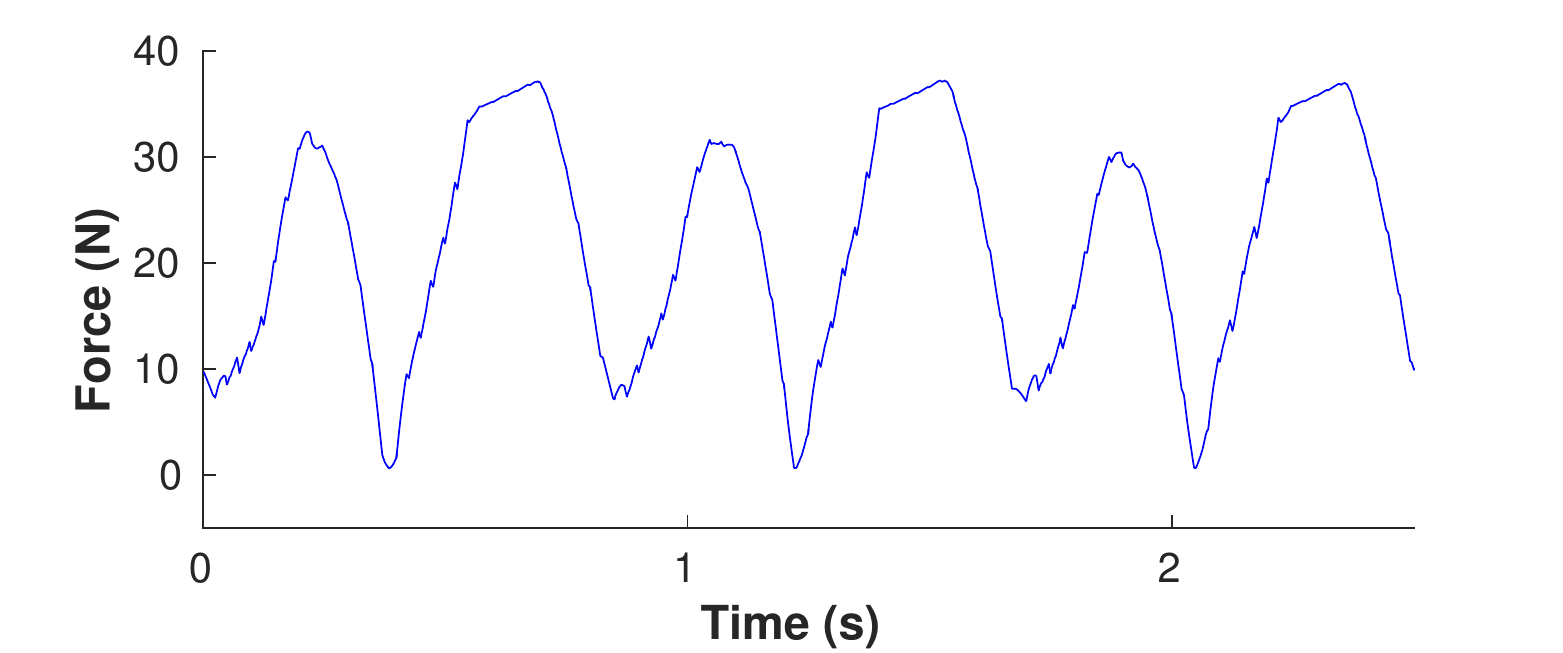}\mylabel{f:single-commanded}{(top)}
    \includegraphics[width=.95\linewidth]{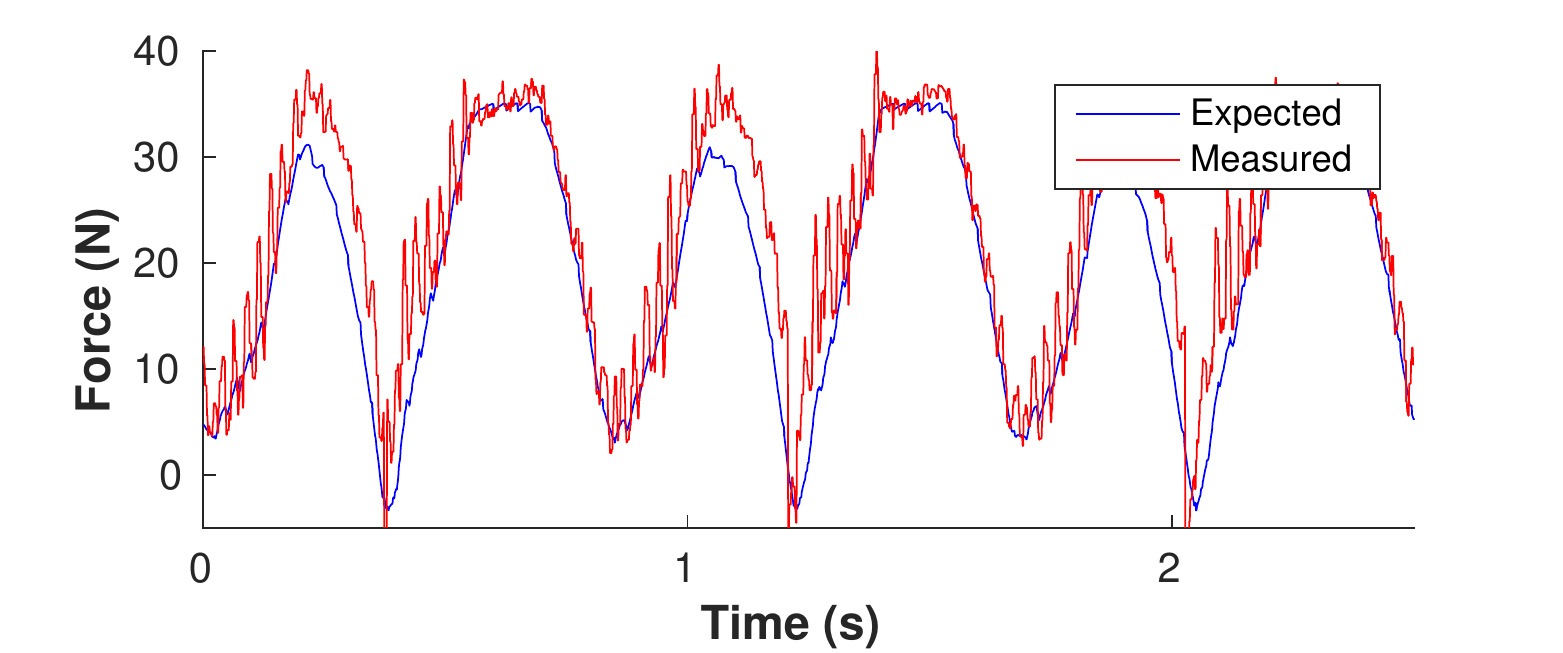}\mylabel{f:single-measured}{(bottom)}\vspace{-2mm}
    \caption{The magnitude of commanded force \ref{f:single-commanded} during circular motion and the corresponding measured force \ref{f:single-measured}.}
    \label{f:single-circular}
\end{figure}

For sanity check, we also compare our expected contact force with the contact force measured using the force/torque sensor in  \fref{single-circular} \ref{f:single-measured}. We can see that the expected contact force are closely aligned with the measured force.

The snapshots shown in \fref{single-exp} \ref{f:single-human} and \ref{f:single-pokes} also show a human subject interacting with the robot by moving, stopping, and poking the end-effector, and the process can be found in the supplementary video.

\subsection{Dual arm holding an object}
\noindent 
In the second experiment, the proposed method is applied on a dual-arm manipulator. In this experiment, we would like to evaluate how well the robot can resist external forces. For this, the task of robot is to hold an object at a static posture while external disturbances are supplied. The robot has no knowledge about the weight of the object nor the magnitude of the external forces.

\begin{figure*}[t!]%
  \centering
   \begin{overpic} [width=.24\linewidth]{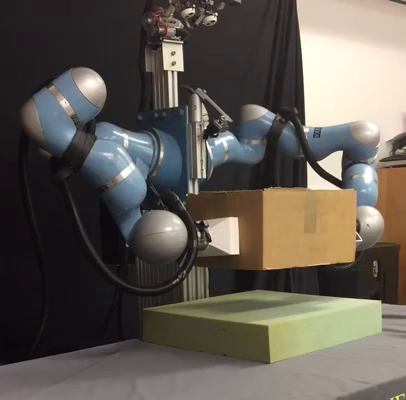}	{\manuallabel{f:dual-static}{(a)}\color{white}\ref{f:dual-static}} 	\end{overpic}~
   \begin{overpic} [width=.24\linewidth]{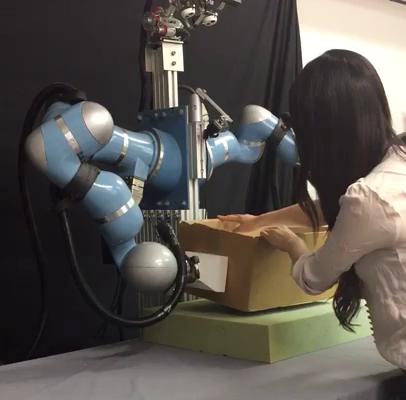}		{\manuallabel{f:dual-push}{(b)}	\color{white}\ref{f:dual-push}}	\end{overpic}~
   \begin{overpic} [width=.24\linewidth]{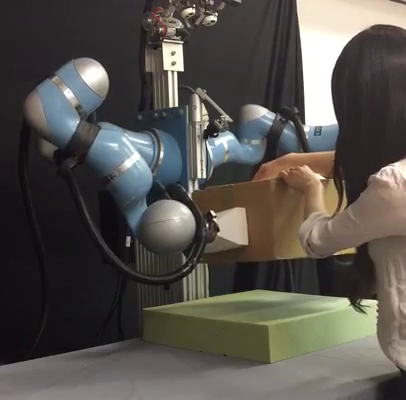}		{\manuallabel{f:dual-up}{(c)}	\color{white}\ref{f:dual-up}}	\end{overpic}~
   \begin{overpic} [width=.24\linewidth]{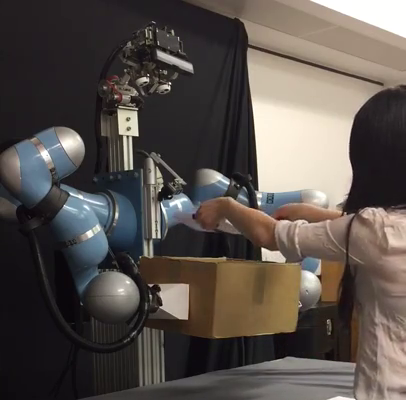}	{\manuallabel{f:dual-weight}{(d)}\color{white}\ref{f:dual-weight}} 	\end{overpic}~  
  \\\vspace{-1mm} 
  \caption{Experiment of a dual-arm robot holding an object. \ref{f:dual-static} The robot holds the box in front of its torso. \ref{f:dual-push} A human pushes the box down, and \ref{f:dual-up} releases. The human adds extra weights on top of the box \ref{f:dual-weight}.}  
  \label{f:dual-hold} 
\end{figure*}

At the beginning of the experiment, as shown in \fref{dual-hold} \ref{f:dual-static}, the robot is holding a rigid box at a position in front of its torso. The size of the box is approximately $20cm\times30cm\times 40cm$ (known to the controller) and the weight is 700 grams (unknown to the controller). A human subject pushes the box about 40 cm downward, stays for a few seconds (\fref{dual-hold} \ref{f:dual-push}), and releases it (\fref{dual-hold} \ref{f:dual-up}). 

This process is repeated a few times, and the norm of the contact force is shown in \fref{dual-force} \ref{f:dual-push-force}, where the colours denote the expected contact force (blue) and the measured contact force (red). Note that the majority of force is due to the end-effectors pushing toward each other. When the robot is at the static position, it squeezes the box with 70 N from both arms. As the person pushes the box down, the robot squeezes the box with a higher force (110 N) to prevent the box from slipping. Note that the robot does not need to measure the external forces in order to know to push harder. The external force is estimated using the displacement of the box relative to its desired position \eref{external-force}.

\begin{figure}[t!]
    \centering  
    \includegraphics[width=\linewidth]{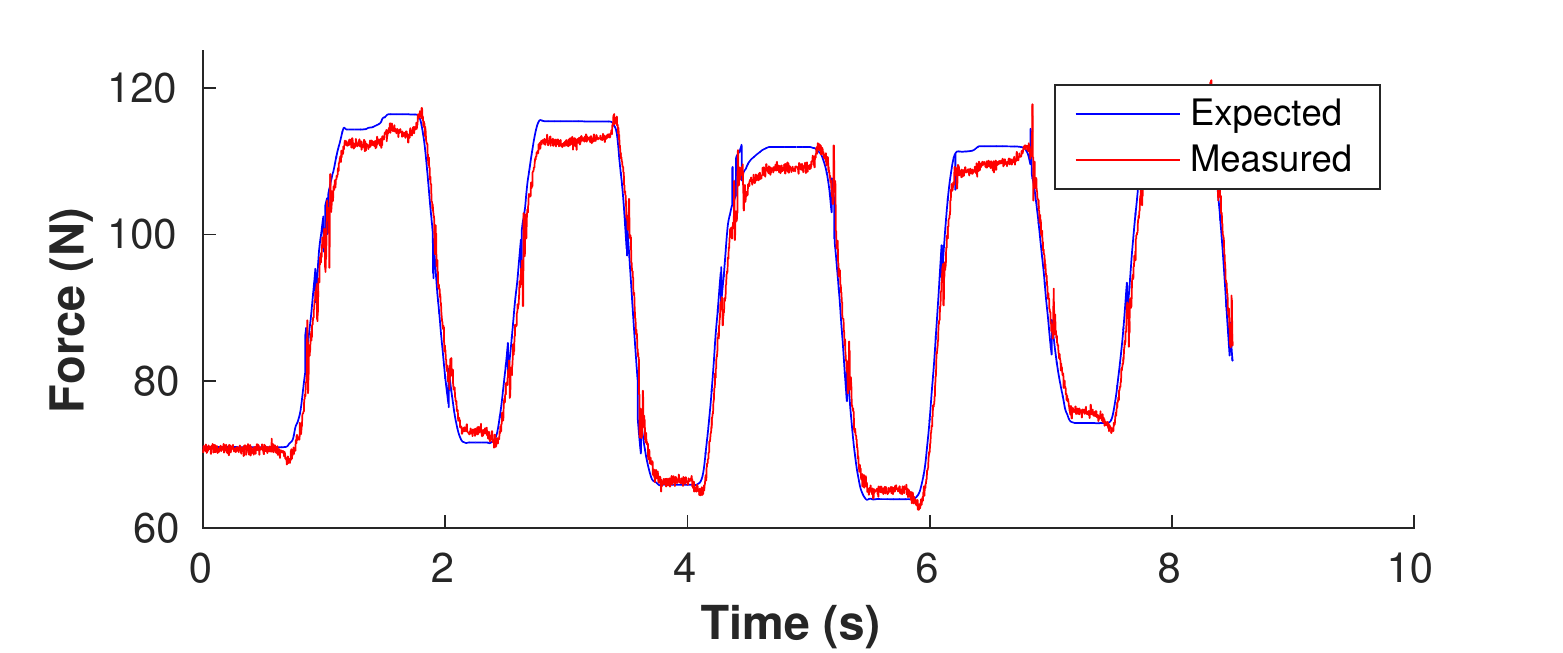}	\mylabel{f:dual-push-force}{(top)}
    \includegraphics[width=\linewidth]{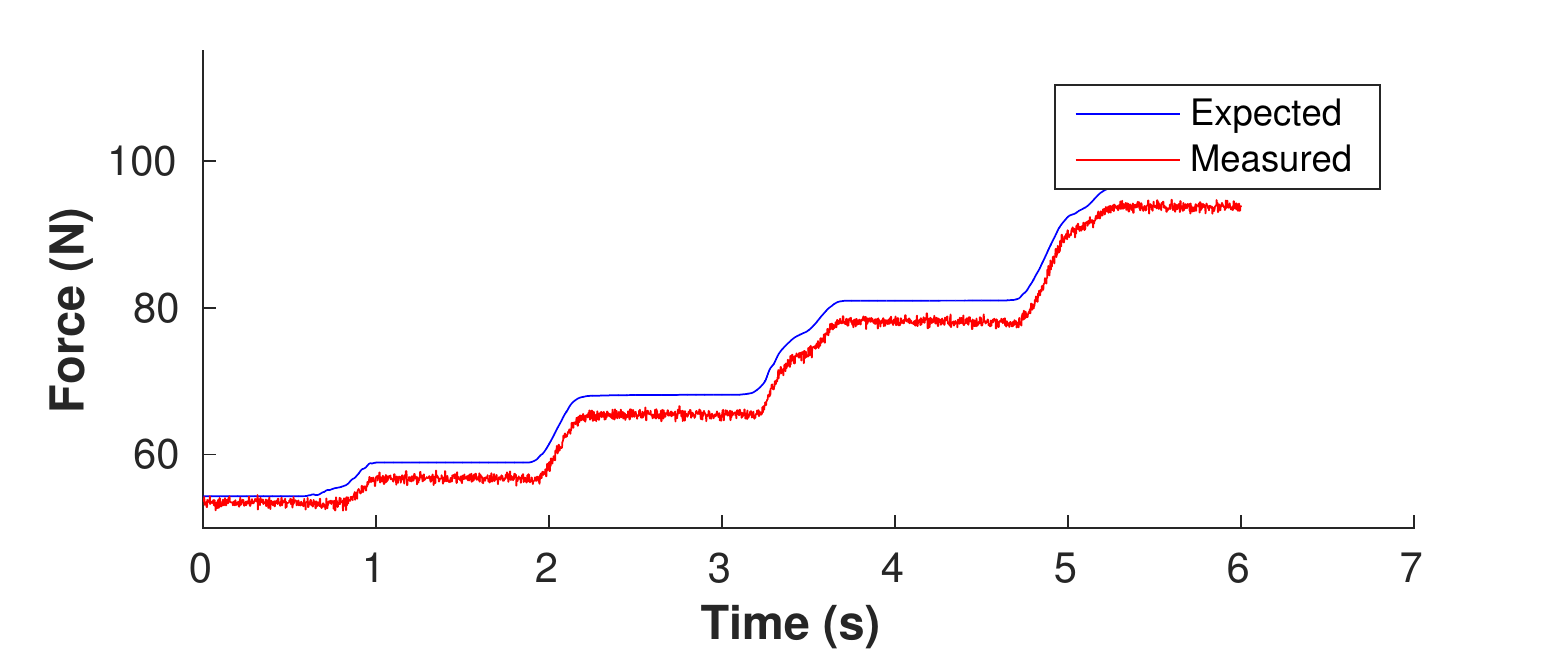}	\mylabel{f:dual-weight-force}{(bottom)}\vspace{-2mm}
    \caption{Norm of the expected contact force (blue) and the measured contact force (red) for the pushing down experiment \ref{f:dual-push-force} and adding weights experiment \ref{f:dual-weight-force}. When external force increases (either by pushing or adding weight), the robot compensates by squeezing harder.}
    \label{f:dual-force}
\end{figure}

In the second half of the experiment, a human subject continuously adds extra weights on top of the box, 500 grams at a time, until a total of 2500 grams are add (see \fref{dual-hold} \ref{f:dual-weight}). The corresponding contact forces are plot in \fref{dual-force} \ref{f:dual-weight-force}. We can clearly see that the contact force increases as the total weight of the object gets heavier.

\subsection{Dual arm manipulating an object}
\noindent 
In our final experiment, we would like to see how well the robot reacts to external forces while performing some task. For this, the robot moves the box in a periodic trajectory, and a human attempts to interrupt the robot by holding the box at a given position (see \fref{dual-traj}).

In this experiment, the trajectory of the box is controlled. The desired trajectory is to follow the circular trajectory in $y,z$-plane, i.e. the desired box position is defined as $\TaskPos_d =\left[0, \radius\cos(\speed t), \radius\sin(\speed t)\right]$

\begin{figure*}[t!]
  \centering
   \begin{overpic} [width=.24\linewidth]{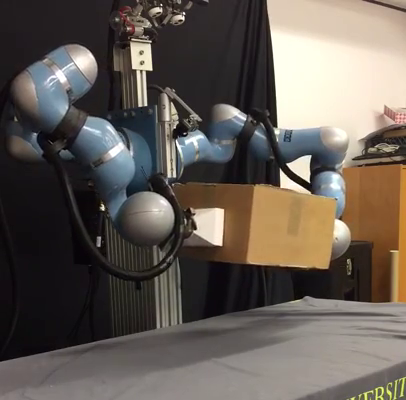}	{\manuallabel{f:dual-traj-1}{(a)}\color{white}\ref{f:dual-traj-1}} 	\end{overpic}~
   \begin{overpic} [width=.24\linewidth]{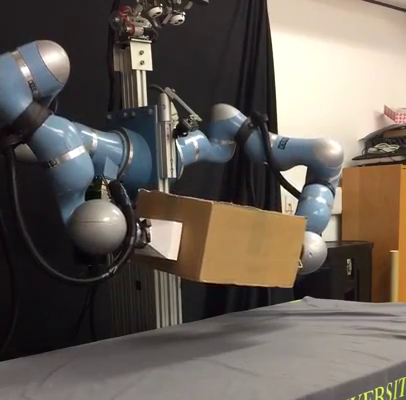}	{\manuallabel{f:dual-traj-2}{(b)}\color{white}\ref{f:dual-traj-2}}	\end{overpic}~
   \begin{overpic} [width=.24\linewidth]{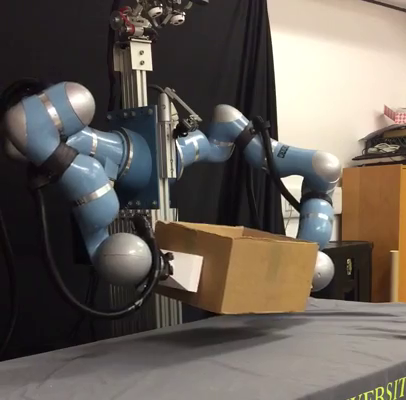}	{\manuallabel{f:dual-traj-3}{(c)}\color{white}\ref{f:dual-traj-3}}	\end{overpic}~
   \begin{overpic} [width=.24\linewidth]{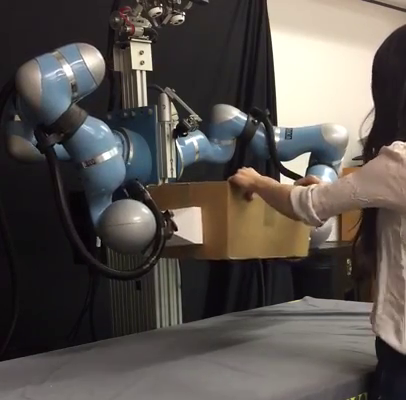}	{\manuallabel{f:dual-traj-human}{(d)}\color{white}\ref{f:dual-traj-human}} 	\end{overpic}~  
  \\\vspace{-1mm} 
  \caption{Experiment of a dual-arm robot moving a box in a circular trajectory. A human attempts to break the trajectory by holding the box.}
  \label{f:dual-traj} 
\end{figure*}

\fref{dual-traj-result} shows the examples of trajectory tracking in $y$-axis \ref{f:dual-traj-y} and $z$-axis \ref{f:dual-traj-z}. The solid lines show the true box positions and the dash lines are the desired box positions. The human tries to hold the box in the middle of the plots, and therefore creates large discrepancies in both $y$ and $z$ axes. Once the human releases the box, the robot continues to follow the circular trajectory. Note we observe that the box positions are consistently lower than desired. Since the mass of the box is unknown to the controller, we do not compensate for its gravity and consequently the true position is always lower than desired. 

\begin{figure}[t!]
    \centering  
    \includegraphics[width=\linewidth]{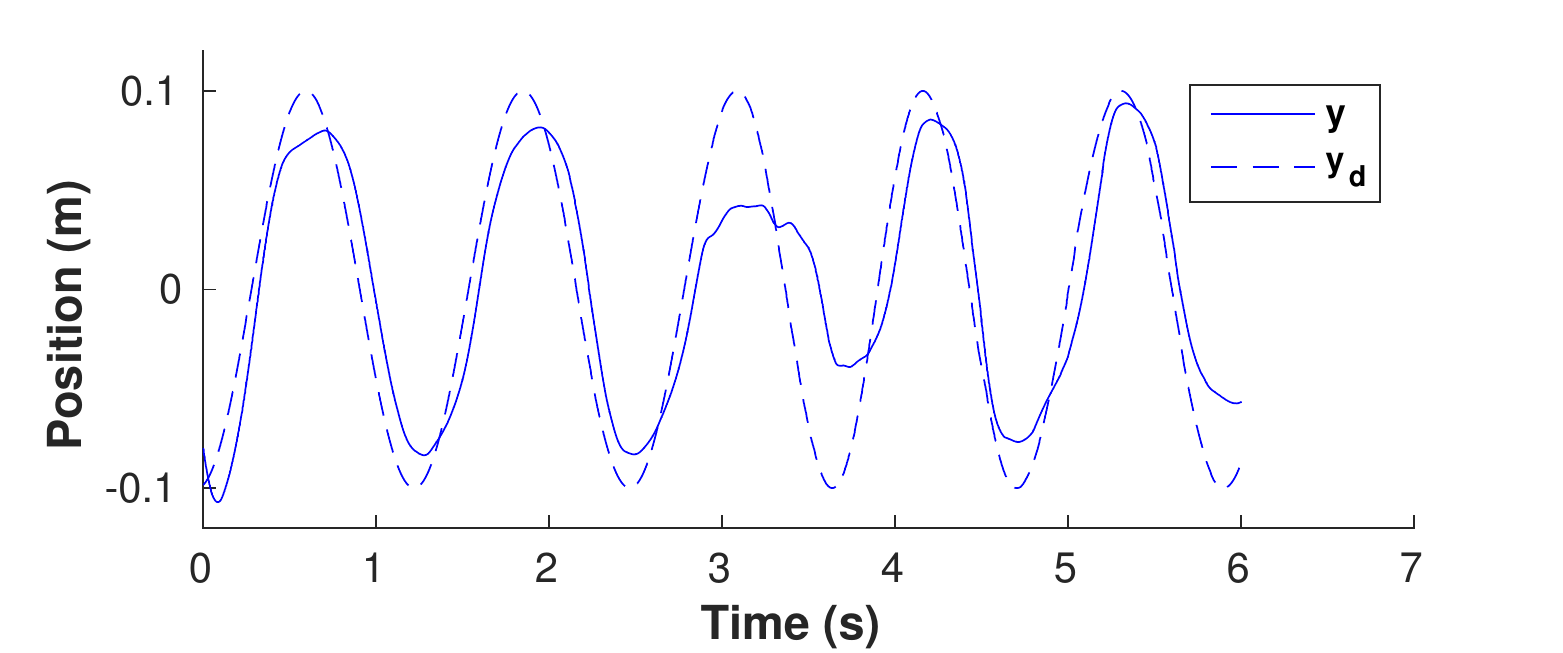}\mylabel{f:dual-traj-y}{(top)}
    \includegraphics[width=\linewidth]{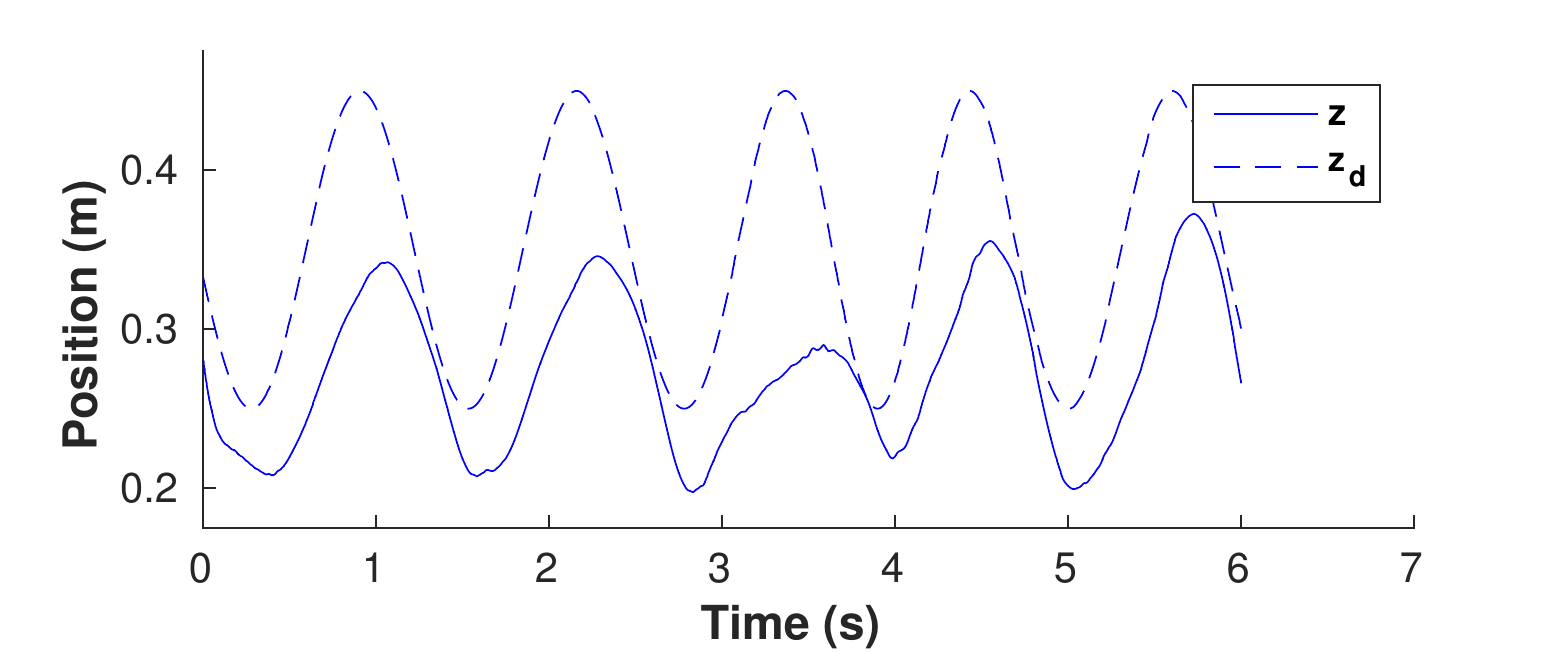}\mylabel{f:dual-traj-z}{(middle)}
    \includegraphics[width=\linewidth]{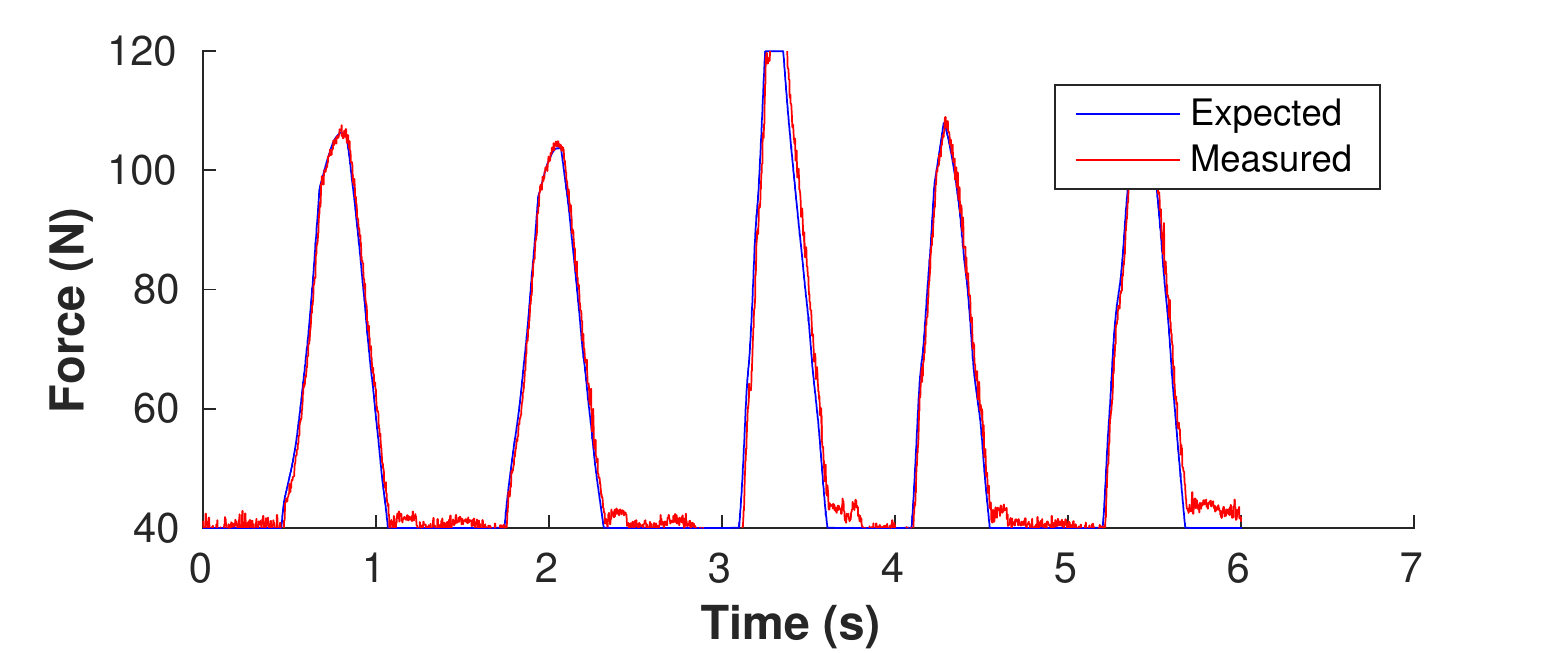}\mylabel{f:dual-traj-force}{(bottom)}
    \vspace{-2mm}
    \caption{A dual-arm robot moves a box in circular trajectory with human interactions. The top two figures show the trajectory tracking in $y$-axis and $z$-axis, and the bottom figure shows magnitude of expected contact force (blue) and measured contact force (red). Human interaction occurs roughly between 3 and 4 seconds.}
    \label{f:dual-traj-result}
\end{figure}

The norm of the applied force is shown in \fref{dual-traj-result} \ref{f:dual-traj-force}. We can see that the force needed to move in a circular motion varies depends on the direction of the motion, i.e.\ when the robot moves downward, the direction of motion is with gravity and hense less force is needed to maintain the grasp.   
\section{Conclusion}    \label{s:conclusion}        	\noindent 
In this paper, a method for dual-arm manipulation with external disturbance is proposed. The problem is formulated in a projected inverse dynamics framework, such that the unconstrained (motion) controller accomplishes the task with desired impedance behaviour, and the constrained component enforces the contact in an optimal manner. The technique is evaluated on both single-arm and dual-arm platforms, showing the proposed method's robustness to unknown disturbances. 

Note that throughout this work, the manipulated object is always assumed to be massless. As a consequence, any inertial or gravity forces due to mass of the object are treated as external disturbances by the impedance controller. The present work demonstrates our controller's robustness and ability to maintain a grasp, subject to unknown human interactions and unknown object inertia.  However, in future work we plan to include estimates of the object's mass/inertia, such that the controller may compensate for these during manipulation.

Furthermore, we have demonstrated that external disturbance forces do not need direct measurement, and may be estimated based on the displacement of the object relative to our desired impedance behaviour. This enables us to to compute optimal constraint forces without direct force measurement. In future work, however, we plan to incorporate F/T contact sensors to allow for inertia shaping in the impedance controller. 

\section*{Acknowledgement} 
\vspace{-1mm}
\noindent This paper was partially supported by the European Commission, within the CogIMon project in the Horizon 2020 Work Programme (ICT-23-2014, grant agreement 644727)

\appendix 
\subsection{Constraint-consistent desired acceleration}
\label{s:joint-acceleration}
\noindent To ensure that the joint accelerations in \eref{dynamic-constrained} is consistent with the desired joint accelerations in \eref{dynamics-external}, we need to solve $\bqddot$ in \eref{dynamics-external}. However, $\P$ is rank deficient, and the term $\P\bM \bqddot$ is not invertible. 

Using the projection matrix,  the constraints in \eref{constraint-jc} can be described as $(\I-\P) \bqdot = 0$. By taking the derivative, $(\I-\P) \bqddot - \dot{\P} \bqdot = 0$, and add this term back to the left hand side of the dynamic equation in \eref{dynamics-external}
\begin{equation*}
  \begin{aligned}
  \P \bM \bqddot + \P \bh &= \P \bM \bqddot + \P \bh + \dot{\P} \bqdot - \dot{\P}\bqdot \\
			&=  \P \bM \bqddot + \P \bh + (\I-\P) \bqddot -  \dot{\P} \bqdot \\
			&=  ( \P \bM + \I-\P ) \bqddot + \P \bh - \dot{\P} \bqdot		
  \end{aligned}
\end{equation*}

\noindent 
Let $\bM_c=\P \bM + \I-\P $, the dynamics equation in \eref{dynamics-unconstrained} can be written as
\begin{equation}
\bM_c \bqddot + \P \bh - \dot{\P} \bqdot = \P \btau +\P\Jacobian_x^\T \ExternalForce 
\label{e:appendix-dynamic-mc}
\end{equation}

\noindent Since $\bM_c$ is invertible, $\bqddot$ can be solved by
\begin{equation}
  \bqddot = \bM_c^{-1} ( \P \btau - \P \bh + \dot{\P} \bqdot + \Jacobian_x^\T \ExternalForce )
  \label{e:dynamic-ddq}
\end{equation}

\subsection{End-effector Force with external disturbance}
\label{s:end-effector-force}
\noindent Multiply \eref{appendix-dynamic-mc} by $\Jacobian_x \bM_c^{-1}$ 
\[ 
  \Jacobian_x \bqddot + \Jacobian_x \bM_c^{-1} ( \P \bh - \dot{\P} \bqdot ) = 
  \Jacobian_x \bM_c^{-1} \P ( \btau + \Jacobian_x^\T \ExternalForce)
\]

\noindent Since $\TaskAcc = \Jacobian_x \bqdot + \dot{\Jacobian}_x \bq$, we replace $\Jacobian_x \bqddot$ with $\TaskAcc - \Jacobian_x \bqdot$ and multiply by $\bLambda_c = (\Jacobian_x \bM_c^{-1} \P \Jacobian_x^\T)^{-1}$ 
\[
\begin{aligned}
\bLambda_c \TaskAcc - \bLambda_c \dot{\Jacobian}_x \bqdot + \bLambda_c \Jacobian_x \bM_c^{-1} (\P\bh - \dot{\P} \bqdot) \\ = \bLambda_c\Jacobian_x \bM^{-1} \P \tau + \ExternalForce
\end{aligned}
\]

\noindent Replacing $\btau$ with $\Jacobian_x^\T  \bF$, and let $\bh_c$ denotes all gravity and velocity terms such that $\bh_c=\bLambda_c \Jacobian_x \bM_c^{-1} (\P\bh - \dot{\P} \bqdot)  - \bLambda_c \dot{\Jacobian}_x \bqdot $,
\[
  \bF + \ExternalForce = \bLambda_c \TaskAcc + \bh_c
\]

\noindent If inertia shaping is avoided (see  \sref{method-impedance} ), the operational space control force is
\begin{equation}
  \bF = \bh_c + \bLambda_c \TaskAcc_d - \bD_d \ErrTaskVel - \bK_d \ErrTaskPos 
  \label{e:appendix-task-force}
\end{equation}

\bibliographystyle{IEEEtran}
\bibliography{bib/IEEEabrv,bib/abbreviations,bib/bibliography}

\begin{thebibliography}{10}
\providecommand{\url}[1]{#1}
\csname url@rmstyle\endcsname
\providecommand{\newblock}{\relax}
\providecommand{\bibinfo}[2]{#2}
\providecommand\BIBentrySTDinterwordspacing{\spaceskip=0pt\relax}
\providecommand\BIBentryALTinterwordstretchfactor{4}
\providecommand\BIBentryALTinterwordspacing{\spaceskip=\fontdimen2\font plus
\BIBentryALTinterwordstretchfactor\fontdimen3\font minus
  \fontdimen4\font\relax}
\providecommand\BIBforeignlanguage[2]{{%
\expandafter\ifx\csname l@#1\endcsname\relax
\typeout{** WARNING: IEEEtran.bst: No hyphenation pattern has been}%
\typeout{** loaded for the language `#1'. Using the pattern for}%
\typeout{** the default language instead.}%
\else
\language=\csname l@#1\endcsname
\fi
#2}}

\bibitem{IROS.2017}
H.-C. Lin, J.~Smith, K.~K. Babarahmati, N.~Dehio, and M.~Mistry, ``A projected
  inverse dynamics approach for dual-arm cartesian impedance control,'' in
  \emph{IEEE International Conference on Intelligent Robots and Systems}, 2017
  (Submitted).

\bibitem{Aghili.2005}
F.~Aghili, ``A unified approach for inverse and direct dynamics of constrained
  multibody systems based on linear projection operator: applications to
  control and simulation,'' \emph{IEEE Transactions on Robotics}, vol.~21,
  no.~5, pp. 834--849, 2005.

\bibitem{Hayati.1986}
S.~Hayati, ``Hybrid position/force control of multi-arm cooperating robots,''
  in \emph{IEEE International Conference on Robotics and Automation},
  vol.~3.\hskip 1em plus 0.5em minus 0.4em\relax IEEE, 1986, pp. 82--89.

\bibitem{Chiaverini.1996}
S.~Chiaverini and B.~Siciliano, ``Direct and inverse kinematics for coordinated
  motion task of a two-manipulator system,'' \emph{Journal of dynamic systems,
  measurement, and control}, vol. 118, pp. 691--697, 1996.

\bibitem{Bonitz.1996}
R.~G. Bonitz and T.~C. Hsia, ``Robust dual-arm manipulation of rigid objects
  via palm grasping-theory and experiments,'' in \emph{IEEE International
  Conference on Robotics and Automation}, vol.~4.\hskip 1em plus 0.5em minus
  0.4em\relax IEEE, 1996, pp. 3047--3054.

\bibitem{Caccavale.2008}
F.~Caccavale, P.~Chiacchio, A.~Marino, and L.~Villani, ``Six-dof impedance
  control of dual-arm cooperative manipulators,'' \emph{IEEE/ASME Transactions
  On Mechatronics}, vol.~13, no.~5, pp. 576--586, 2008.

\bibitem{Hirche.2015}
S.~Erhart and S.~Hirche, ``Internal force analysis and load distribution for
  cooperative multi-robot manipulation,'' \emph{IEEE Transactions on Robotics},
  vol.~31, no.~5, pp. 1238--1243, 2015.

\bibitem{Righetti.2013}
L.~Righetti, J.~Buchli, M.~Mistry, M.~Kalakrishnan, and S.~Schaal, ``Optimal
  distribution of contact forces with inverse-dynamics control,'' \emph{Int.
  Journal of Robotics Research}, vol.~32, no.~3, pp. 280--298, 2013.

\bibitem{Lee.20016}
Y.~Lee, S.~Hwang, and J.~Park, ``Balancing of humanoid robot using contact
  force/moment control by task-oriented whole body control framework,''
  \emph{Autonomous Robots}, vol.~40, no.~3, pp. 457--472, 2016.

\bibitem{Bicchi.2000}
A.~Bicchi and V.~Kumar, ``Robotic grasping and contact: A review,'' in
  \emph{IEEE International Conference on Robotics and Automation},
  vol.~1.\hskip 1em plus 0.5em minus 0.4em\relax IEEE, 2000, pp. 348--353.

\bibitem{Kerr1986-rh}
J.~Kerr and B.~Roth, ``\BIBforeignlanguage{en}{Analysis of multifingered
  hands},'' \emph{\BIBforeignlanguage{en}{Int. J. Rob. Res.}}, vol.~4, no.~4,
  1986.

\bibitem{Buss.1996}
M.~Buss, H.~Hashimoto, and J.~B. Moore, ``Dextrous hand grasping force
  optimization,'' \emph{IEEE Transactions on Robotics and Automation}, vol.~12,
  no.~3, pp. 406--418, 1996.

\bibitem{Trinkle.1997}
J.~C. Trinkle, J.-S. Pang, S.~Sudarsky, and G.~Lo, ``On dynamic
  multi-rigid-body contact problems with coulomb friction,'' \emph{ZAMM-Journal
  of Applied Mathematics and Mechanics/Zeitschrift f{\"u}r Angewandte
  Mathematik und Mechanik}, vol.~77, no.~4, pp. 267--279, 1997.

\bibitem{Hogan.1985}
N.~Hogan, ``Impedance control: An approach to manipulation, part i - theory,''
  \emph{ASME Journal of Dynamic Systems, Measurement, and Control}, vol. 107,
  pp. 1--7, 1985.

\bibitem{Schneider.1992}
S.~A. Schneider and R.~H. Cannon, ``Object impedance control for cooperative
  manipulation: Theory and experimental results,'' \emph{IEEE Transactions on
  Robotics and Automation}, vol.~8, no.~3, pp. 383--394, 1992.

\bibitem{1987.IJRR.Khatib}
O.~Khatib, ``A unified approach for motion and force control of robot
  manipulators: The operational space formulation,'' \emph{IEEE J. Robotics \&
  Automation}, vol.~3, no.~1, pp. 43--53, 1987.

\bibitem{Mistry.2011}
M.~Mistry and L.~Righetti, ``Operational space control of constrained and
  underactuated systems,'' \emph{Robotics: Science and systems}, 2011.

\bibitem{Ott.2008}
C.~Ott, \emph{Cartesian impedance control of redundant and flexible-joint
  robots}.\hskip 1em plus 0.5em minus 0.4em\relax Springer, 2008.

\bibitem{Udwadia.2001}
R.~Kalaba and F.~Udwadia, ``Analytical dynamics with constraint forces that do
  work in virtual displacements,'' no.~2, pp. 211--217, 2001.

\bibitem{Murray.1994}
R.~M. Murray, Z.~Li, and S.~Sastry, \emph{A mathematical introduction to
  robotic manipulation}.\hskip 1em plus 0.5em minus 0.4em\relax CRC press,
  1994.

\bibitem{Boyd.2007}
S.~P. Boyd and B.~Wegbreit, ``Fast computation of optimal contact forces,''
  \emph{IEEE Transactions on Robotics}, vol.~23, no.~6, pp. 1117--1132, 2007.

\bibitem{Aghili.2016}
F.~Aghili and C.-Y. Su, ``Control of constrained robots subject to unilateral
  contacts and friction cone constraints,'' in \emph{IEEE International
  Conference on Robotics and Automation}.\hskip 1em plus 0.5em minus
  0.4em\relax IEEE, 2016, pp. 2347--2352.

\bibitem{GUROBI}
{Gurobi Optimization}, ``Gurobi optimizer reference manual,''
  http://www.gurobi.com.

\end{thebibliography}

\end{document}